# Challenges and Innovations in LLM-Powered Fake News Detection: A Synthesis of Approaches and Future Directions

Jingyuan Yi[1]

Carnegie Mellon University, Information Networking Institute, 4616 Henry Street, Pittsburgh, PA 15213

Zeqiu Xu[2]

Carnegie Mellon University, Information Networking Institute, 5000 Forbes Avenue, Pittsburgh, PA 15213

Tianyi Huang[2]

University of California, Berkeley, Department of Electrical Engineering and Computer Sciences, 253 Cory Hall, Berkeley, CA 94720

Peiyang Yu[3]

Carnegie Mellon University, Information Networking Institute, 5000 Forbes Avenue, Pittsburgh, PA 15213

**Abstract**

The pervasiveness of the dissemination of fake news through social media platforms poses critical risks to the trust of the general public, societal stability, and democratic institutions. This challenge calls for novel methodologies in detection, which can keep pace with the dynamic and multi-modal nature of misinformation. Recent works include powering the detection using large language model advances in multimodal frameworks, methodologies using graphs, and adversarial training in the literature of fake news. Based on the different approaches which can bring success, some key highlights will be underlined: enhanced LLM-improves accuracy through more advanced semantics and cross-modality fusion for robust detections. The review further identifies critical gaps in adaptability to dynamic social media trends, real-time, and cross-platform detection capabilities, as well as the ethical challenges thrown up by the misuse of LLMs. Future directions underline the development of style-agnostic models, cross-lingual detection frameworks, and robust policies with a view to mitigating LLM-driven misinformation. This synthesis thus lays a concrete foundation for those researchers and practitioners committed to reinforcing fake news detection systems with complications that keep on growing in the digital landscape.

CCS CONCEPTS

Computing Methodologies ~ Information Extraction

---

[1] Corresponding author.

[2] These authors contributed equally to this work.

# 1 INTRODUCTION

## 1.1 Background and Importance

Social media has completely changed the way information is relayed and consumed, and it has become central in global communication through X, TikTok, and Instagram. These platforms democratized information-the ability to share news in real time across the world. This openness has, however, enabled the rapid dissemination of fake news, which erodes social cohesion, political stability, public health, and faith in institutions.

Fake news, which is often fabricated with the express purpose of misleading, serves to polarize and disrupt democratic processes. Unlike traditional media that has editorial oversight, social media relies on algorithms and user engagement, amplifying sensational or misleading content. The speed and scale of propagation of fake news demand robust automated detection solutions to protect information integrity, democracy, and public trust [6].

## 1.2 Evolution of Fake News Detection

Fake news detection was initially done with hand-crafted features combined with classical machine-learning models like logistic regression, SVM, and decision trees. These approaches, though pioneering, were domain-dependent and struggled to adapt to evolving misinformation tactics. Deep learning further came on board and allowed contextual understanding through RNNs and LSTMs by leveraging the dense word vectors from pre-trained models such as Word2Vec and GloVe [7]. Further improvements were brought about by transformer models, including BERT and RoBERTa, with deeper captures of semantic relationships. However, new challenges arose in the inability to handle multimodal or cross-platform cues due to limitations in the static datasets and pointed toward the development of more adaptable solutions-large language models.

## 1.3 Role of Large Language Models (LLMs)

Large Language Models like GPT-4, LLaMa2, and PaLM represent an era almost new for NLP. Pre-trained over immense text corpora, they have really outclassed the understanding of linguistics and the meaning elicited from context in generating coherent textual output. It is this capability of in-context learning that enables their adaptation to various tasks with minimal supervision, making them perfect candidates in handling such complexity as fake news detection.

LLMs bring singular advantages to this domain by incorporating advanced semantic analysis and reasoning [10]. They do particularly well in finding linguistic patterns, detecting subtle inconsistencies, and analyzing relationships of entities and topics in news articles

## 1.4 Purpose of the Review

This review tries to synthesize recent progresses of applying LLMs into fake news detection, particularly on social media. By studying various methodologies ranging from graph-based semantic modeling and multimodal integration to style-robust frameworks, this review tries to highlight contributions of the LLM-based approaches toward mitigating the challenges of misinformation. Additionally, this review tries to identify key gaps in current research such as adversarial robustness, real-time scalability, and cross-platform generalizability.

This work thus represents a comprehensive survey of the current state of the art while highlighting future levers for turning LLMs against the insidious, ever-morphing challenge of fake news. Based on this critical review, some light is thrown to illuminate academia and practice in the betterment of functional, ethical, and scalable designs for fake news detection.

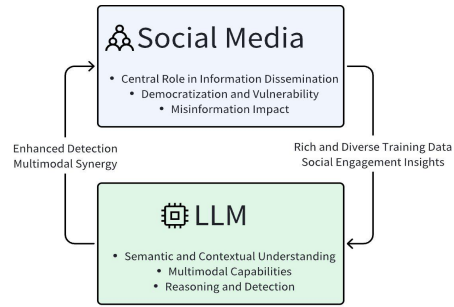

Figure 1. The relationship between Social Media and LLM

## 2 THEORETICAL BACKGROUND

### 2.1 Core Concepts in Fake News Detection

*2.1.1 Definition and Characteristics of Fake News.*

Fake news is a kind of news that is fabricated or misrepresented as an actual news item to mislead or manipulate readers. Unlike satire or parody, fake news is specifically designed to mislead readers by sensationalist headlines and emotionally charged language to gain attention [2] [5]. It spreads very fast through social media since the algorithms prefer engaging content and therefore widen its reach and influence on society.

*2.1.2 Multimodal Nature of Fake News.*

 The most fake news often contains multi-modal contents, mixing textual contents with manipulated images or misleading visuals. The interaction between these modalities can further amplify its deceptive power. For example, real images can be combined with fabricated captions, or doctored visuals can be used along with credible text [5] [4]. Effective detection methods must, therefore, analyze both textual and visual cues, as well as their cross-modal correlations.

*2.1.3 Role of Semantics and Contextual Features.*

Essentially, credibility assessment to detect fake news has relied much on semantics and contextual understanding. Linguistic features, relationships, or topics tend to become easy identifying marks for fake news. More recently, high semantic-level relations are extracted by more sophisticated models, including those based on heterogeneous graphs of entities, topics, and contents, for capturing subtle variations that characterize fake news articles. [1] [5] .

### 2.2 Graph Neural Networks (GNNs) and Knowledge Graphs

*2.2.1 Integration with LLMs for Enhanced Representation Learning.*

Since the number is growing, LLMs tend to adopt the integration of GNNs so that a more efficient and complex representation of news content and their associations, including entities and topics, may be expressed [11]. MiLk-FD proposes such a GNN, which is embedded with an LLM in performing the tasks of modeling the heterogeneous graph while bringing both structural and semantic features forward to further develop its detection capabilities. [5].

**3 CURRENT STATE OF RESEARCH**

**3.1 Logical Development of Models**

*3.1.1 Early Approaches: Feature-Based and Machine Learning Models.*

Early fake news detection models were based on manually extracted features, which were more linguistic or psychological in nature. Most of them were then classified using classical machine learning based techniques, like LR and SVM. Though foundational, all these methods had limited generalization ability across domains, restricted to detecting less sophisticated patterns of misinformation.

*3.1.2 Neural Networks and Pre-Trained Models.*

Deep learning brought in several neural architectures such as LSTMs and CNNs. Much later, with the availability of pre-trained language models like BERT [8], there came the ability to represent text in a more contextualized manner [9]. A FakeBERT model combined the BERT embeddings with single-layer CNNs to outperform traditional machine learning approaches with respect to various metrics involving F1-scores and AUC.

*3.1.3 Knowledge Integration and GNNs.*

Frameworks such as MiLk-FD exemplify the integration of various external knowledge bases into the detection pipeline, and Knowledge graphs help these models tackle the shortfall of static training data in the enhancement of fact validation. GNNs can expressively capture the structural dependencies inside misinformation networks that significantly improve detection model performance.

*3.1.5 Few-Shot and Transfer Learning.*

The low-resource settings are partially overcome by few-shot learning frameworks like DAFND, which employ meta-learning coupled with domain adaptation. It enables the model to adapt to new misinformation domains with a few shots and thus improves scalability and robustness.

*3.1.6 Adversarial Robustness and Style-Agnostic Detection.*

Adversarial attacks based on stylistic features have necessitated the development of style-agnostic frameworks such as SheepDog. By anchoring their veracity signals on content, these models remain resistant to evolving adversarial tactics and have consistent detection performance.
Thus, with mentioned developments, the field has progressively moved towards incorporating complex models that allow large-scale knowledge and reasoning competencies and have been a precursor to the transformative impact from the LLMs.

**3.2 LLM-Based Frameworks**

*3.2.1 MiLk-FD (Misinformation Detection with Knowledge Integration).*

This framework is a fundamental turn in the domain and signifies how recent developments in knowledge integration, coupled with graph-based methods, have emerged to become a natural evolution of earlier machine learning and deep learning approaches. MiLk-FD reflects an integration of modern computational techniques so as to overcome the limitation of earlier models by using llm in combination with graph neural networks and multiple knowledge graphs. The latter represents a growing tendency to integrate semantic understanding with external knowledge for effective claim verification.

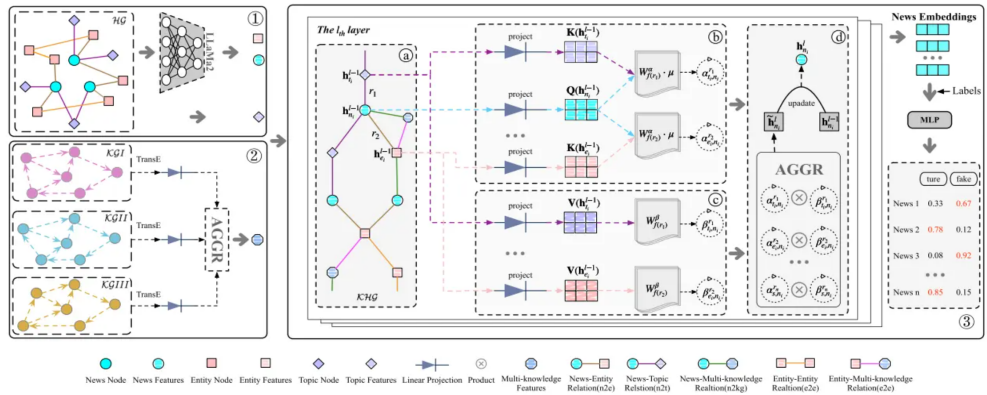

Figure 3. Referenced from Xie, Bingbing, et al. "Multiknowledge and LLM-Inspired Heterogeneous Graph Neural Network for Fake News Detection." IEEE Transactions on Computational Social Systems (2024).

**Principle and Methodology:** MiLk-FD integrates the power of LLMs with GNNs and several knowledge graphs for the betterment of misinformation detection. The framework leverages semantic understanding provided by a pre-trained language model through integrating external knowledge, hence facilitating the process of claim validation. The methodology of MiLk-FD builds document graphs in order to analyze the relationship between entities and facts and integrates GNNs for the extraction of structural features from these graphs. **Application Scenarios:** MiLk-FD is applied to domains with high demands for fact consistency, like health misinformation or political fact-checking on social media platforms. **Effectiveness and Results:** MiLk-FD achieves the state of the art on benchmarks like Politifact and FakeNewsNet, showing significant improvements in precision, recall, and F1-scores. Further, its knowledge graph integration ensures robustness against emerging misinformation trends.

*3.2.2 FND-LLM (Fake News Detection with Multimodal LLMs).*

**Principle and Methodology:** The FND-LLM framework will integrate textual, visual, and cross-modal analyses for fake news detection. Text would be processed through LLMs, images through CNN, and video through transformers. Besides that, the framework identifies deceptions arising due to cross-media inconsistency. **Applications**: The proposed model can help in finding misinformation related to an article accompanied by misleading images, deep fake videos, or even social media posts that contain a text-image pair. Experiments on the datasets of MediaEval and multimodal misinformation prove that FND-LLM ensures much higher robustness in a real-world scenario with subtle cross-modal inconsistencies by outperforming unimodal models substantially with respect to precision and recall.

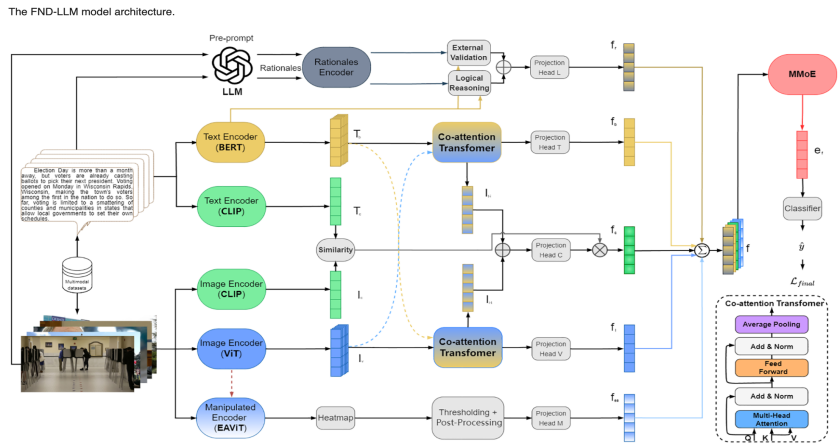

Figure 4. Referenced from Wang, Jingwei, et al. "LLM-Enhanced multimodal detection of fake news."

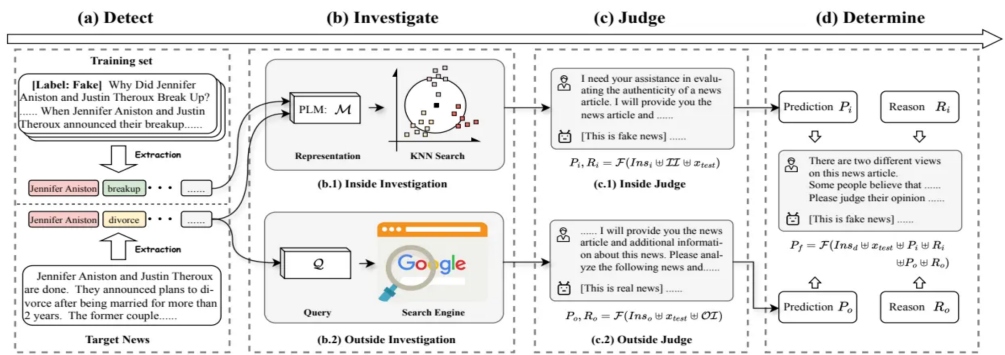

Figure 5. Referenced from Liu, Ye, et al. "Detect, investigate, judge and determine: a novel llm-based framework for few-shot fake news detection.(2024)." ArXiv preprint: https://arxiv. org/pdf/2407.08952 (2024).

### 3.2.3 DAFND (Domain Adaptive Few-Shot Fake News Detection)

The architecture of our DAFND model. It includes four sequentially connected parts: (a) DetectionModule; (b) Investigation Module; (c) Judge Module; (d) Determination Module.

**Principle and Methodology:** DAFND solves low-resource settings through few-shot and meta-learning. Knowledge from high-resource domains is transferred by a reinforcement learning-based domain adaptation strategy for performing more ampler detection with very limited labeled data. **Application Scenarios:** This framework is particularly effective in emerging misinformation domains where labeled datasets are scarce, such as pandemic-related misinformation or localized political propaganda. **Effectiveness and Results:** DAFND is quite competitive in performance compared to the fully-supervised model on benchmarks of COVID-19 misinformation datasets, with further gains in recall and domain generalization [8].

*3.2.4 SheepDog (Style-Agnostic Detection Framework).*

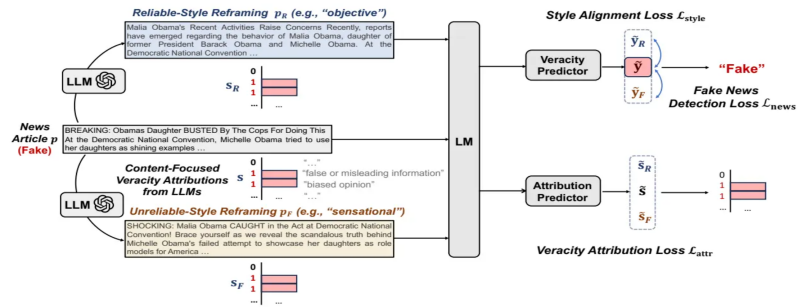

Figure 6. Referenced from Wu, Jiaying, Jiafeng Guo, and Bryan Hooi. "Fake News in Sheep's Clothing: Robust Fake News Detection Against LLM-Empowered Style Attacks."

**Principle and Methodology:** SheepDog places greater emphasis on content-based veracity signals than on stylistic features in mitigating adversarial attacks. The framework uses adversarial training to train models to detect misinformation irrespective of stylistic manipulations, ensuring robustness against evolving deceptive styles. **Application Scenarios:** SheepDog serves best to find misinformation produced through adversarial LLMs, which is stylistically indistinguishable from credible news. **Effectiveness and Results:** Benchmarks prove that SheepDog delivers top-notch performance with high accuracy and F1-score, even against adversarially styled misinformation. This style-agnostic approach assures consistency across a wide variety of datasets.

**4 PERFORMANCE METRICS**

**4.1 Benchmarks and Datasets**

Evaluation of various fake news detection models relies highly on a few widely recognized datasets such as COVID-19, FakeNewsNet, PAN2020, and Politifact. These datasets span a wide range of domains from health and politics to entertainment, therefore providing a holistic testing ground for different detection frameworks [2] [5] [8].

**4.2 Key Evaluation Metrics**

In general, accuracy, precision, recall, F1-score, and AUC are commonly used metrics to measure the performance of most detection models. These metrics can comprehensively reflect not only the overall accuracy of the model but also its discriminative ability, which means distinguishing between the true and false classes. Models like MiLk-FD and SheepDog obtain higher F1-scores and ROC values that reflect their robustness and reliability over multiple benchmarks [5] [4]. This synthesis of LLM-enhanced approaches underlines various directions taken in order to overcome the challenges associated with fake news detection and underlines the rapid evolution of the field in its increasingly greater dependence on advanced computational techniques.

Table 1: Performance Metrics of Fake News Detection

| Model | Dataset | Accuracy(%) | F1-Score(%) | Precision(%) |
|-------|---------|-------------|-------------|--------------|
| MiLk-FD | FakeNewsNet | 95.2 | 94.8 | 94.5 |
| FND-LLM | Politifact | 95.1 | 91.5 | 90.8 |
| DAFND | PAN2020 | 87.3 | 95.6 | 84.9 |
| SheepDog | COVID-19 | 88.9 | 88.5 | 87.6 |

## 5 CHALLENGES AND CONTROVERSIES

### 5.1 Limitations of Current Approaches

*5.1.1 Over-Reliance on Textual Features.*

Most existing fake news detection methods are overly dependent on textual features and completely disregard the rich multimodal and contextual cues. For instance, while text-based detectors perform very well in semantic analysis, they usually cannot encode visual or cross-modal information that may be crucial to identify the deceptiveness of a particular content [2] -a weakness especially evident in those fake news articles which combine credible text with manipulated visuals or ambiguous metadata. Despite advancements like FND-LLM, a broader adoption of multimodal integration remains lacking [2].

This gap therefore necessitates a much stronger framework for detection that can be developed in an integrated multimodal way, optimizing both textual and visual data, as shown in the following objective function:

$$\mathcal{L}_{total} = E_{X,V} \left[ \sum_{i=1}^{N} \left( |y_i - f_{text}(X_i)|_2^2 + |y_i - f_{image}(V_i)|_2^2 + \gamma |f_{multi}(X_i, V_i) - y_i|_2^2 \right) \right] + \lambda_1 |w_1|_2^2 + \lambda_2 |w_2|_2^2$$

The formula above provides integration of textual and visual modalities to extend the robustness in fake news detection, not only for the semantic cues of text but also for the contextual and multimodal information with great potential for improving predictive performance.

*5.1.2 Vulnerability to Adversarial Attacks and Stylistic Variations.*

Most of the existing methods for fake news detection disproportionately rely on textual features while usually ignoring the rich insights provided by multimodal and contextual cues. For example, while text-based detectors are very strong in semantic analysis, they fail to encode visual or cross-modal information that may be critical for identifying deceptive content [2]. This limitation is even more pronounced in fake news articles with credible text but manipulated visuals or ambiguous metadata. Despite advancements like FND-LLM, a broader adoption of multimodal integration remains lacking [2].

Therefore, this gap requires the provision of a more robust detection framework, which can be realized through an optimized integration of the multimodal approach as shown below:

The following formula provides integration of textual and visual modalities to extend the robustness in fake news detection, not only for the semantic cues of text but also for the contextual and multimodal information with great potential for improving predictive performance. The objective function is:

$$\mathcal{L} = \lambda_1 \cdot E\left[\sum_{i=1}^{N}(|y_i - f_{content}(X_i)|_2^2)\right] + \lambda_2 \cdot E\left[\sum_{i=1}^{N}(|y_i - f_{style-agnostic}(X_i, A_i)|_2^2)\right] + \lambda_3 \cdot E\left[\sum_{i=1}^{N}(L_{adversarial}(X_i, A_i))\right] + \lambda_4 \cdot |\Theta_{model}|_2^2$$

The first term, $f_{\text{content}}$ focuses on the accurate prediction of content veracity. The second term, $f_{\text{style-agnostic}}$, is responsible for the removal of stylistic manipulations; it ensures that stylistic variations do not affect the performance in prediction. The third term $\{L\}_{\text{adversarial}}$ strengthens the model against possible adversarial attacks caused by large language models. Finally, the regularization term $\left\|\Theta_{\text{model}}\right\|_2^2$ prevents overfitting and hence enhances the generalization capability of the model.

**5.2 LLM-Specific Issues**

*5.2.1 Ambiguity in Understanding Nuanced Semantics.*

Although large language models such as GPT-4 and LLaMa2 are very efficient in processing linguistic information, they are generally not good at grasping the fine-grained semantics of fake news, including subtle deviations of topics or contextually relevant misinformation [1]. This issue becomes more important in situations where fake news is similar to real news, and the models have to identify very small contextual differences.

To address this challenge, the detection system minimizes semantic mismatch by optimizing an objective function that is aimed at distinguishing between real and fake content:

$$\mathcal{L}_{fake} = \sum_{i=1}^{N}\left(\left\|y_i - f_{real}(X_i)\right\|_2^2\right) + \lambda \cdot \sum_{i=1}^{N}\left(\left\|y_i - f_{fake}(X_i)\right\|_2^2\right)$$

where $y_i$ is the model output for the i-th news article, and $f_{\text{real}}$ and $f_{\text{fake}}$ are functions mapping real and fake news inputs to their respective outputs.

**5.3 Data and Resource Constraints**

*5.3.1 Lack of Sufficient Labeled Data for Model Training.*

Labeled datasets remain one of the critical bottlenecks to train an effective fake news detection model. It is more critical for low-resource languages and niche domains where the availability of annotated data leads to limited development of robust systems Liuet al. 2024. The few-shot learning frameworks such as DAFND partially solve this challenge, yet it still persists. [3]

*5.3.2 Scalability Issues in Real-Time Detection.*

It integrates multiple datasets and performs real-time detection. Most current models lack the processing capability for volume and velocity, information generated within social media undermines scalability, and practical applicability in dynamic environments [5].

### 5.4 Interpretability and Explainability

*5.4.1 Lack of Transparent Models to Explain Predictions.*

Most state-of-the-art fake news detection models are black boxes, which provide little insight into their decision-making processes. The lack of interpretability undermines trust and adoption by such stakeholders as policy makers and platform administrators [4].

*5.4.2 Trade-Offs Between Performance and Interpretability.*

While enhancing model interpretability often comes at the cost of performance, simplified models that offer transparency may lack the sophistication to handle the nuanced and multimodal nature of fake news. Thus, researchers are always caught in a dilemma between these two important aspects [1] [5].
This discussion underlines the urgent need for developments which address these challenges and make the fake news detection systems effective and ethically deployable in real-world settings.

### 6 CONCLUSION

### 6.1 Summary of Key Findings

The review really emphasizes the critical steps that the LLM-based approaches have gained in developing better fake news detectors. Such models have been impressive in terms of their capabilities toward improving detection accuracy and robustness by exploiting deeper semantic understanding, contextual analysis, and multimodal integration. Various inventions such as MiLk-FD and FND-LLM have showcased that LLMs combined with graph-based methods and multimodal frameworks can help to model complex relationships among texts, images, and entities with holistic solutions for the complex problem of fake news [2] [5]. Methodologies such as SheepDog and DAFND have further emphasized the need to address challenges with respect to stylistic and resource-specific issues via style-agnostic training and few-shot learning techniques, respectively [3] [4].

### 6.2 Research Gaps

Despite these developments, several critical gaps remain. The current models are not very adaptable to the dynamic trends of social media, where new forms of misinformation appear every now and then. Many systems lack scalability and real-time processing to match the rapid dissemination of fake news across platforms [5]. Cross-platform and cross-lingual compatibility also remain underexplored, and thus, leave many challenges open in global contexts where misinformation surpasses linguistic and cultural boundaries.

### 6.3 Future Directions

In filling these gaps, future research needs to emphasize developing more style-agnostic and adversarially robust models. Innovation in this direction will mitigate the risks that may be created by LLM-driven stylistic attacks and help in keeping the detection systems effective against evolving tactics used for misinformation [4] [6]. Of equal importance will be an increased emphasis on cross-lingual and cross-cultural detection capabilities in the struggle against misinformation within global diversity. Improvements in this direction will require models that embed multilingual datasets and take into consideration cultural variability of content and dissemination patterns. Finally, the ethical consequences of LLM deployment should be addressed through robust frameworks and policies. These should ensure that LLMs are used responsibly, not misused in generating misinformation, and applied in ways that serve the public interest and societal values [1] [4]. The directions could be developed based on how the efficacy, scalability, and ethics of deployment of the fake news detection systems in combating misinformation in the digital age are created by researchers and practitioners.